\documentclass[lettersize,journal]{IEEEtran}
\usepackage{amsmath,amsfonts}
\usepackage{algorithmic}
\usepackage{algorithm}
\usepackage{array}
\usepackage[caption=false,font=footnotesize,labelfont=rm,textfont=rm]{subfig}
\usepackage{textcomp}
\usepackage{stfloats}
\usepackage{url}
\usepackage{verbatim}
\usepackage{graphicx}
\usepackage{cite}
\usepackage{booktabs}
\usepackage{multirow}
\usepackage{threeparttable}
\usepackage{color} 
\usepackage{hyperref}
\hypersetup{
     colorlinks = true,
     linkcolor = green,
     filecolor = green,
     citecolor = blue,      
     urlcolor = blue,
     }

\begin{document}

\title{TPLLM: A Traffic Prediction Framework Based on Pretrained Large Language Models}

\author{Yilong Ren, Yue Chen, Shuai Liu, Boyue Wang, Haiyang Yu, and Zhiyong Cui
\thanks{This work was supported by the NSFC (52202378); in part by the Open Research Project Program of the State Key Laboratory of Internet of Things for Smart City (SKL-IoTSC(UM)-2021-2023/ORP/GA08/2022), the Youth Talent Support Program of Beihang University under Grant (YWF-22-L-1239), and the Ministry of Transport of PRC Key Laboratory of Transport Industry of Comprehensive Transportation Theory (MTF2023002). \emph{(Corresponding author: Zhiyong Cui.)}}
\thanks{Yilong Ren, Yue Chen, Shuai Liu, Haiyang Yu and Zhiyong Cui are with the State Key Laboratory of Intelligent Transportation Systems, School of Transportation Science and Engineering, Beihang University, Beijing 100191, China (e-mail: yilongren@buaa.edu.cn; cyu6369@buaa.edu.cn; shuailiu@buaa.edu.cn; hyyu@buaa.edu.cn; zhiyongc@buaa.edu.cn).}
\thanks{Boyue Wang is with the Faculty of Information Technology,
Beijing University of Technology, Beijing 100124, China (e-mail: wby@bjut.edu.cn).}
}



\maketitle

\begin{abstract}
Traffic prediction constitutes a pivotal facet within the purview of Intelligent Transportation Systems (ITS), and the attainment of highly precise predictions holds profound significance for efficacious traffic management. The precision of prevailing deep learning-driven traffic prediction models typically sees an upward trend with a rise in the volume of training data. However, the procurement of comprehensive spatiotemporal datasets for traffic is often fraught with challenges, primarily stemming from the substantial costs associated with data collection and retention. Consequently, developing a model that can achieve accurate predictions and good generalization ability in areas with limited historical traffic data is a challenging problem. It is noteworthy that the rapidly advancing pretrained Large Language Models (LLMs) of recent years have demonstrated exceptional proficiency in cross-modality knowledge transfer and few-shot learning. Recognizing the sequential nature of traffic data, similar to language, we introduce TPLLM, a novel traffic prediction framework leveraging LLMs. In this framework, we construct a sequence embedding layer based on Convolutional Neural Networks (CNNs) and a graph embedding layer based on Graph Convolutional Networks (GCNs) to extract sequence features and spatial features, respectively. These are subsequently integrated to form inputs that are suitable for LLMs. A Low-Rank Adaptation (LoRA) fine-tuning approach is applied to TPLLM, thereby facilitating efficient learning and minimizing computational demands. Experiments on two real-world datasets demonstrate that TPLLM exhibits commendable performance in both full-sample and few-shot prediction scenarios, effectively supporting the development of ITS in regions with scarce historical traffic data. 
\end{abstract}


\section{Introduction}
\IEEEPARstart{T}{o} mitigate the mounting strain of traffic congestion and curb the economic losses and environmental pollution it spawns, numerous countries have embarked on fostering the development and implementation of Intelligent Transportation Systems (ITS). Traffic prediction is a core functionality of ITS \cite{yin2021deep}, and the attainment of precise predictive outcomes is critically important for both traffic status analysis and effective traffic management. For example, accurate traffic flow prediction empowers traffic management authorities to issue timely congestion alerts, thereby enabling drivers to circumvent congested routes, which directly contributes to a reduction in average vehicle travel time and, consequently, a decrease in greenhouse gas emissions \cite{zhao2022smart}.

\begin{figure}[!t]
\centering
\includegraphics[width=0.8\linewidth]{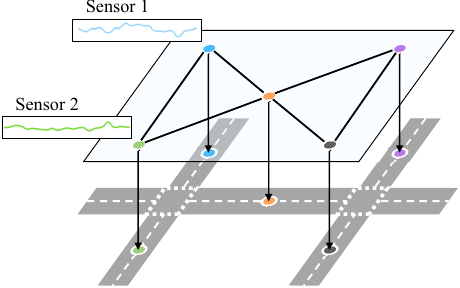}
\caption{Correlation between time-series traffic data.}
\label{fig_1}
\end{figure}

With the technological advances in recent years, emerging technologies spearheaded by deep learning have furnished increasingly efficient and accurate support to a multitude of ITS functionalities. Spatio-temporal data of traffic presents a robust foundation for deep learning-driven traffic prediction methodologies. This category of data embodies intricate spatio-temporal characteristics and is mainly constituted by time-series data collected by multiple sensors, coupled with the corresponding spatial information of the underlying road network. According to the spatial structure of the road network, there is a correlation between each time-series traffic data \cite{cao2021survey}, \cite{yuan2021survey}, as shown in Fig.~\ref{fig_1}.

Existing deep learning-based traffic prediction models usually extract spatio-temporal features from traffic data through multiple extractors, the performance of these models usually increases with the amount of training data \cite{emmert2020introductory}. To ensure good accuracy, most traffic prediction models require datasets containing longer history data for training. However, due to the high cost of long-term data collection and storage, there are difficulties in constructing a comprehensive traffic spatio-temporal dataset in most regions. This constraint thereby limits the widespread application of certain data-driven traffic prediction models. Furthermore, these models are often trained only for specific tasks, which may lead to overfitting and thereby degrade the generalization ability of the model. In summary, it is still challenging to develop models that are resilient to overfitting and capable of delivering accurate predictions in areas with limited historical traffic data.

\IEEEpubidadjcol

\begin{figure}[!t]
\centering
\includegraphics[width=\linewidth]{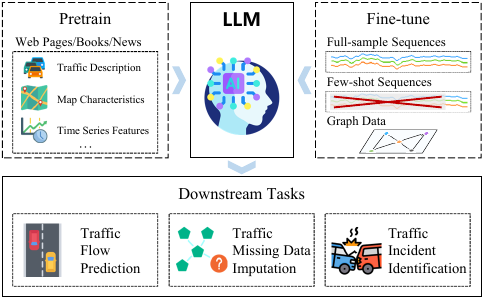}
\caption{Pretrained LLM for traffic tasks.}
\label{fig_2}
\end{figure}

Pretrained Large Language Models (LLMs), which have rapidly emerged in recent years, offer a promising solution to the aforementioned challenge. Pretrained LLMs are deep learning models trained on large-scale high-quality generalized datasets to capture universal patterns and information. LLMs are widely recognized for generative tasks due to their capabilities of powerful few-shot learning \cite{brown2020language} and cross-modality knowledge transfer \cite{lu2021pretrained}. Endowed with an extensive array of parameters and a wealth of pre-existing knowledge, LLMs have found applications across a diverse range of domains, notably including transportation. These models exhibit remarkable potential for swift adaptation to a variety of downstream tasks, such as traffic prediction, data imputation, and incident identification. This adaptability is facilitated through the process of fine-tuning, which requires only minimal data \cite{houlsby2019parameter} to significantly extend the models' capabilities, as depicted in Fig.~\ref{fig_2}. The fine-tuning mechanism leverages a considerable number of pretrained parameters, which are kept frozen to prevent overfitting, thereby enhancing the models' ability to generalize across different tasks and datasets.


\begin{figure}[!t]
\centering
\includegraphics[width=0.9\linewidth]{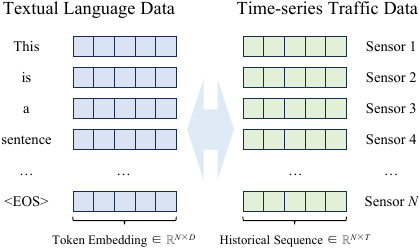}
\caption{Similarity between time-series traffic data and natural language.}
\label{fig_3}
\end{figure}

Although LLMs are extensively utilized across various fields, they were initially devised for processing natural language \cite{zhao2023survey} through a token embedding mechanism, which seems unsuitable for the time-series data typically encountered in transportation applications. Nevertheless, Fig.~\ref{fig_3} illuminates a significant structural similarity between multivariate time-series traffic data and textual data, with both being representable as collections of vectors of consistent dimensionality. This congruence effectively narrows the divide between these distinct types of data, unveiling a promising path for applying LLMs to the analysis of traffic data. Inspired by this insight, we are motivated to pursue innovative modifications of LLMs, aiming to harness their potential for analyzing traffic data and deciphering complex spatiotemporal patterns.

In order to introduce pretrained LLMs to the traffic prediction task and overcome the few-shot challenge caused by data cost, we propose TPLLM, a framework for traffic prediction based on pretrained LLMs. The central idea of the TPLLM is to shape the multivariate time-series traffic data into a form that is understandable by LLMs in a token embedding-like manner, thus exploiting the prior knowledge in the LLMs. To further enhance the model's understanding of the spatial features of the traffic data, we also append graph-structured spatial information of the road network to the input. The final output from the LLMs is used in order to generate traffic prediction results. To optimize training efficiency and fine-tuning effectiveness, we employ a Parameter-Efficient Fine-Tuning (PEFT) approach, specifically Low-Rank Adaptation (LoRA) \cite{hu2021low}, significantly reducing training costs without compromising performance. Our experiments show that with the powerful prior knowledge and inference capabilities of LLMs, the TPLLM can efficiently complete the regular traffic prediction task and perform equally well on the few-shot prediction task. 

In summary, our contributions can be listed as follows:

\begin{itemize}
\item{We propose TPLLM, a framework for traffic prediction based on pretrained LLMs, to cope with full-sample and few-shot traffic prediction tasks.}
\item{An embedding module is designed based on convolutional neural networks and graph convolutional networks. This module aims to enable LLMs to understand time-series data and to fuse spatiotemporal features implied within the traffic data.}
\item{To reduce  training costs and maintain high fine-tuning quality, we applied a cost-effective fine-tuning method, LoRA, to TPLLM. This approach preserves TPLLM's existing knowledge, also improving its efficiency in making accurate traffic predictions with limited data.}
\item{We conducted experiments in scenarios with sufficient and limited training data, respectively, to make the proposed method highly relevant to the realistic situations. The results validate that the prior knowledge in pretrained LLMs can be effectively applied to traffic prediction.}
\end{itemize}

The paper is structured as follows: Section 2 gives an overview of related studies. Section 3 explains the traffic prediction issue and the structure of our proposed framework. Section 4 covers the experimental results on real-world data. Section 5 concludes the paper and suggests areas for future research.

\section{Related Work}
In this section, the related work of this paper is reviewed, including research on traffic prediction and LLMs for time series prediction.
\subsection{Traffic Prediction}
Traffic data exhibit strong dynamic correlations in both spatial and temporal dimensions, making traffic prediction a challenging task. Early work on traffic prediction generally relied on statistical methods or conventional machine learning methods such as Autoregressive Integrated Moving Average (ARIMA) \cite{hamed1995short}, Support Vector Machine (SVM) \cite{ding2002traffic}, and K-Nearest Neighbor (KNN) \cite{zheng2014short}. These methods view traffic data as a simple time series, which makes it difficult to capture the nonlinear spatio-temporal features in the data, and therefore have limitations and low prediction accuracy.

In recent years, with the rapid development of computer technology, deep learning methods, especially graph-based methods, have been widely used in traffic prediction. In this field, Recurrent Neural Networks (RNNs) \cite{elman1991distributed} and its variants Long Short-Term Memory (LSTM) \cite{hochreiter1997long} and Gate Recurrent Unit (GRU) \cite{cho2014learning} are often used to extract the temporal dependence of traffic data. Meanwhile, Graph Convolutional Networks (GCNs) \cite{bruna2013spectral} are often used to extract the spatial dependence of traffic data. Convolutional Neural Networks (CNNs) \cite{lecun1989backpropagation} and attention mechanisms \cite{vaswani2017attention} can also be integrated to identify salient information. Several traffic prediction methods have achieved commendable outcomes by integrating multiple deep learning strategies. For instance, STGCN \cite{yu2017spatio} consists of ST-Conv blocks that capture spatio-temporal correlations through GCNs and CNNs in each block. ASTGCN \cite{guo2019attention} employs a spatio-temporal attention mechanism that combines GCNs and CNNs, respectively, allowing the model to dynamically learn the correlation between space and time. STSGCN \cite{song2020spatial} connects graph-structured data at different time steps into a single graph, which can directly and simultaneously capture local spatial-temporal correlations. Despite these achievements, these methods usually focus on tasks with standard-sized training data, which means they require a large amount of historical data for training to achieve good accuracy. This reliance on extensive datasets presents a challenge when dealing with limited historical traffic data scenarios.

\subsection{Pretrained LLMs}
In recent years, propelled by the evolution of computational devices and the emergence of vast text corpora, a multitude of Transformer-based pretrained LLMs have demonstrated remarkable capabilities in tackling diverse natural language processing tasks \cite{zhao2023survey}. Researchers found that model performance could be improved by stacking modules, so they further investigated the scaling effect by increasing the parameter scales to larger size. When the parameter size exceeds a certain level, these language models not only achieve significant performance improvements, but also exhibit some unique capabilities that are not available in small-scale models, such as the few-shot learning capability \cite{brown2020language}.

OpenAI proposed Generative Pre-Trained Transformer (GPT) \cite{radford2018improving} based on the Transformer decoder, which shows a strong capability by training on a large amount of corpus data. The basic principle of GPT is to compress various types of knowledge into a decoder-only Transformer model through linguistic modeling so that the knowledge can be memorized and act as a general-purpose task solver. Based on this idea, GPT-2 \cite{radford2019language}, GPT-3 \cite{brown2020language}, and GPT-4 \cite{achiam2023gpt} with increasing number of parameters have been gradually released, and they not only perform well in a variety of NLP tasks, but also show very good performance on some specialized tasks that require domain adaptability. General Language Model (GLM) \cite{du2021glm} is the first open-source LLM optimized for Chinese-English bilingual training. The main architecture of GLM consists of a stack of Transformer decoders which apply an autoregressive blank infilling technique for self-supervised training for a wide range of tasks. Large Language Model Meta AI (LLaMA) \cite{touvron2023llama} is an open and efficient base LLM released by Meta AI. LLaMA normalizes the inputs of each Transformer layer instead of the outputs, and removes the absolute position embedding, replacing it with the rotational position embedding at each layer. Bidirectional Encoder Representations from Transformers (BERT) \cite{devlin2018bert} is a language model based on Transformer encoders. It cannot perform exact generation but can reconstruct the original data using bi-directional contextual information and is therefore commonly used for content understanding tasks.

We believe that the nature of the time series prediction task is a generative task on the predicted sequences. Therefore, generative LLMs based on Transformer decoders may be more applicable to our study, such as GPT.

\subsection{LLMs for Time-series Prediction}
Due to the exceptional few-shot learning capability \cite{brown2020language} and cross-modality knowledge transfer proficiency \cite{lu2021pretrained} of LLMs, their applications can be expanded into numerous scenarios across different domains through the PEFT technology without necessitating complete retraining.

However, there are relatively few studies applying pretrained LLMs to traffic prediction, and the main research focuses on the field of general time series prediction. Zhou et al. \cite{zhou2024one} proposed a generalized time-series analysis framework based on cross-modality knowledge migration of pretrained LLMs. This is the first time that a pretrained LLM is used for time series analysis tasks including prediction, classification, interpolation, and anomaly detection. The framework makes input embedding and positional embedding of the input time series and applies the PEFT method to the LLM. Following this, a variety of generalized time-series processing frameworks based on pretrained LLMs have emerged. For example, Chang et al. \cite{chang2023llm4ts} proposed a framework based on a two-stage fine-tuning LLM. Firstly, the model is aligned with the time-series characteristics through supervised fine-tuning, followed by further fine-tuning guided by the downstream task. As another example, Rasul et al. \cite{rasul2023lag} first applied LLaMA as a time-series prediction base model and demonstrated its good few-shot learning ability through experiments with 20\% to 80\% of the training data. In addition, some studies have focused on tokenization of time series to generate embeddings that are more manageable by LLM. For instance, Sun et al. \cite{sun2023test} designed a time-series encoder based on comparative learning to obtain applicable embeddings; Liu et al. \cite{liu2024autotimes} aligned time series segments with language tokens to projects each segment into the embedding space of LLMs.

In the field of traffic time-series processing with application of pretrained LLMs, there are only a few preliminary studies. Chen et al. \cite{chen2023gatgpt} first used pretrained LLMs for a traffic spatio-temporal task, where they discerned spatial dependencies for data imputation through the graph attention mechanism. Liu et al. \cite{liu2024spatial} first used pretrained LLMs for traffic prediction by learning spatial location and global temporal representations of tokens through a spatio-temporal embedding module. However, this method did not take into account the graph-structured spatial features of traffic road networks.

Inspired by the above studies and considering the complex spatio-temporal characteristics of traffic data, we construct a novel framework dedicated to traffic prediction based on pretrained LLMs.

\section{Methodology}
In this section, the traffic prediction task is first described. Subsequently the components of the TPLLM are described in detail.
\subsection{Task description}
The data used in traffic prediction tasks can be described as features of a graph, and traffic data detected by a single sensor can be considered as graph node features. Therefore, we use the graph $G = (V, E, \mathbf{A})$ to define the road network in our study, where $V$ is a finite set consisting of $N$ traffic sensor nodes; $E$ is a finite set consisting of edges between connected nodes; and $\mathbf{A} \in \mathbb{R}^{N \times N}$ is the adjacency matrix of the graph $G$, which denotes the connectivity between nodes. In graph $G$, each sensor node samples the traffic data with the same frequency. We use $\mathbf{x}_t = \{x_1, ..., x_n, ..., x_N\} \in \mathbb{R}^{N}$ to denote the temporal characteristics of the graph $G$ at time $t$, where $x_n$ denotes the characteristic data, such as flow and average velocity, collected by a single sensor $n$ at time $t$.

\begin{figure*}[!t]
\centering
\includegraphics[width=\linewidth]{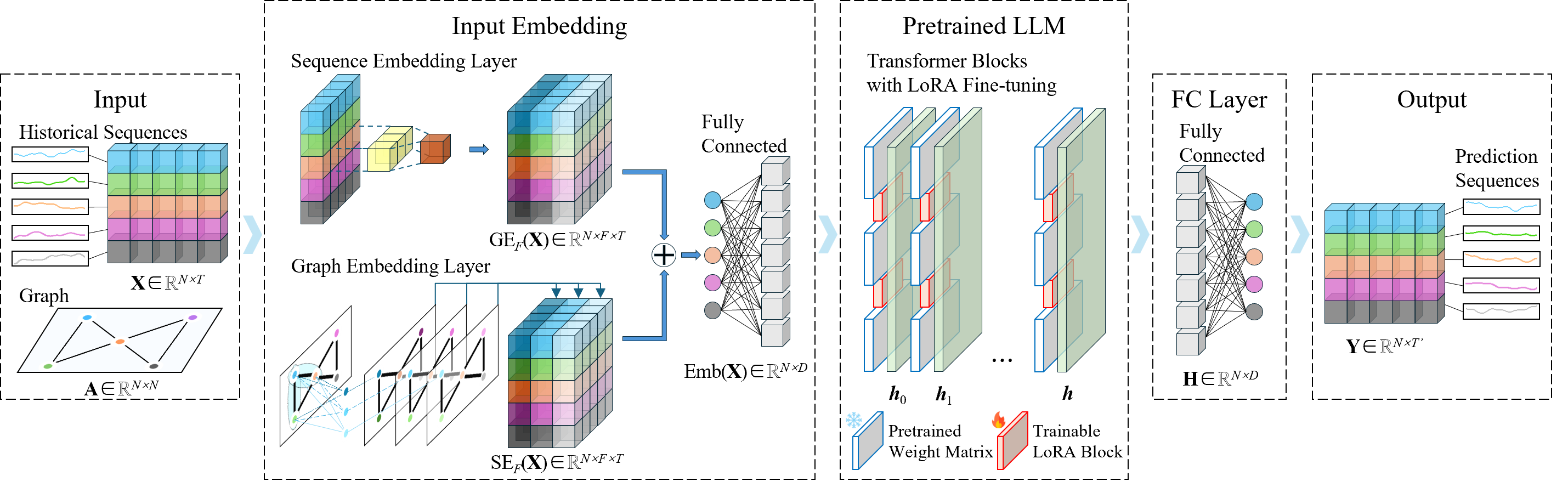}
\caption{Structure of the TPLLM.}
\label{fig_4}
\end{figure*}

The aim of traffic prediction tasks is to find a model $f(.)$. The input of the model is a historical sequence of the last $T$ time steps $\mathbf{X} = \{\mathbf{x}_{t-T+1}, ..., \mathbf{x}_t\} \in \mathbb{R}^{N \times T}$ and the adjacency matrix $\mathbf{A}$ of the graph $G$. The output of the model is a prediction sequence of the next $T'$ time steps $\mathbf{Y} = \{\mathbf{y}_{t+1}, …, \mathbf{y}_{t+i}, ..., \mathbf{y}_{t+T'}\} \in \mathbb{R}^{N \times T'}$, where $\mathbf{y}_{t+i} \in \mathbb{R}^{N}$ is the prediction value of the feature of the graph $G$ at the i-th time step in the future. The prediction process of the model is as follows:

\begin{equation}
\label{deqn_ex1}
\mathbf{Y} = f(\mathbf{X}, \mathbf{A}).
\end{equation}

Based on the deep learning theory, we can approximate $f(.)$ using the approximator $f_{\theta}(.)$ based on a deep learning model, where $\theta$ is learnable parameters. In this study, we aim to construct $f_{\theta}(.)$ based on a pretrained LLM to utilize the few-shot learning capability of the LLM to make the model suitable for regular and small-sample traffic prediction tasks.

\subsection{Overview of the TPLLM}
The structure of the TPLLM is shown in Fig.~\ref{fig_4}. We use a pretrained Transformer-based LLM as a backbone for analyzing and predicting traffic spatio-temporal data. In order to enable the pretrained LLM to process the time-series data while taking into account the spatial features in the data, we designed an input embedding module that consists of two components:

\begin{itemize}
\item{Sequence Embedding Layer: This layer uses a CNN to process the sequential traffic data, extracting the temporal dependencies and patterns.}
\item{Graph Embedding Layer: This layer uses a GCN to process the adjacency matrix of the road network, extracting the spatial dependencies and patterns.}
\end{itemize}

Subsequently, in the pretrained LLM, we use LoRA \cite{hu2021low} in each transformer block to fine-tune the model to achieve good performance at a small training cost. Finally, the prediction results are output through a linear layer.

\subsection{Input Embedding}
In order to make the pretrained LLM adaptable to the traffic prediction task, the spatio-temporal data needs to be input in a form that can be understood by the LLM. Based on the similarity between time-series traffic data and natural language, we consider the historical data sequence of a single sensor during a period $T$ as a word, and the data $\mathbf{X}$ of all sensors in the road network during this period as a sentence.

Sensor nodes in a road network do not have sequential relationships among them like words in a sentence, but rather positional relationships with a graph structure. GCNs are feature extractors for graph-structured data that extract features from the aggregation of node adjacency information through the input data and adjacency matrix. Therefore, we designed a graph embedding layer based on GCNs to extract spatial information from road networks. The graph embedding $GE_{F}(\mathbf{X}) \in \mathbb{R}^{N \times F \times T}$ with $F$ channels is computed as follows:

\begin{equation}
\label{deqn_ex2}
\tilde{\mathbf{A}} = \mathbf{A} + \mathbf{I},
\end{equation}

\begin{equation}
\label{deqn_ex3}
GE_{F}(\mathbf{X}) = ReLU(\tilde{\mathbf{D}}^{-\frac{1}{2}}\tilde{\mathbf{A}}\tilde{\mathbf{D}}^{-\frac{1}{2}}\mathbf{X}\mathbf{W} + \mathbf{b}),
\end{equation}

\noindent where $\mathbf{I}$ denotes the unit matrix, $\tilde{\mathbf{D}}$ denotes the degree matrix of $\tilde{\mathbf{A}}$, $\mathbf{W}$ denotes the learnable parameter matrix, $\mathbf{b}$ denotes the learnable bias vector, and $ReLU(.)$ is the ReLU activation function.

In addition, we design a sequence embedding layer based on 1-D CNNs to extract features from the time series themselves. The sequence embedding $SE_{F}(\mathbf{X}) \in \mathbb{R}^{N \times F \times T}$ with $F$ channels is computed as follows:

\begin{equation}
\label{deqn_ex4}
SE_{F}(\mathbf{X}) = Conv1d_{F}(\mathbf{X}),
\end{equation}

\noindent where $Conv1d_{F}(.)$ denotes a 1-D convolution operation with $F$ filters.

Finally, the input needs to be compatible with the selected pretrained LLM. The graph embedding is fused with the sequence embedding, followed by a linear layer that maps the temporal dimension of the fused information to the LLM’s embedding size $D$ to obtain $\mathbf{M} \in \mathbb{R}^{N \times F \times D}$. Finally, the state of the last feature channel of $\mathbf{M}$ is taken to get the input of the LLM. The input embedding $Emb(\mathbf{X}) \in \mathbb{R}^{N \times D}$ is computed as follows:

\begin{equation}
\label{deqn_ex5}
\mathbf{M} = Linear_{D}(LN(ReLU(GE_{F}(\mathbf{X}) + SE_{F}(\mathbf{X})))),
\end{equation}

\begin{equation}
\label{deqn_ex6}
Emb(\mathbf{X}) = \{m_{iFj} \vert 0 \leq i < N, 0 \leq j < D \},
\end{equation}

\noindent where $LN(.)$ denotes a layer normalization, $Linear_{D}(.)$ denotes a linear layer that maps the input to dimension $D$, $m_{iFj} \in \mathbf{M}$ denotes the elements in $\mathbf{M}$.

\subsection{Pretrained LLM and PEFT}
Despite the similarity between traffic data and textual data, as generic pretrained LLMs are still dedicated to textual data, we should not make them directly handle the embedding of traffic spatio-temporal data without adjustment. Although retraining the entire LLM may solve this problem, the required computational resources are too large to be acceptable.Based on the above, we need to apply the PEFT method to adjust the pretrained LLM with lower computational resources to make it suitable for downstream traffic prediction tasks.

It is demonstrated that pretrained LLMs that freeze most of the parameters can, applying the PEFT method, be comparable to models that are fully trained on downstream tasks \cite{lu2021pretrained}. The PEFT of pretrained LLMs aims to improve the performance of pretrained LLMs on new tasks by adjusting or introducing trainable parameters to minimize the number of trainable parameters and computational complexity.

\begin{figure}[!t]
\centering
\includegraphics[width=\linewidth]{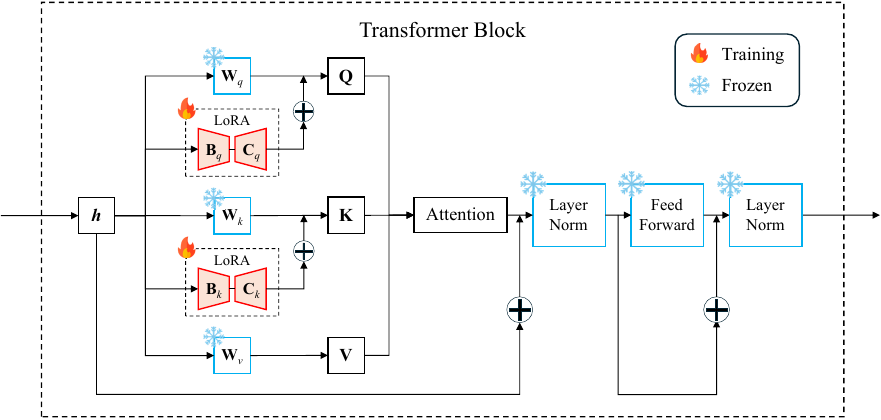}
\caption{Structure of the Transformer blocks.}
\label{fig_5}
\end{figure}

We use LoRA \cite{hu2021low}, a PEFT method that injects trainable rank decomposition matrix $\mathbf{B}$ and $\mathbf{C}$ into each Transformer block in the LLM to significantly reduce the size of trainable parameters. The structure of the Transformer blocks in the LLM is shown in Fig.~\ref{fig_5}.

To avoid introducing additional noise to the model, $\mathbf{B}$ is zero initialized and $\mathbf{C}$ is Gaussian initialized. LoRA is applied to the Query and Key of the attention layers in the LLM. We denote the rank of the LoRA module by $r$, which is a hyperparameter reflecting the amount of information the module can contain. Assuming that the input of Query and Key in an attention layer is $\mathbf{h}_{0}$ with size $d$ and the output is $\mathbf{h}$ with size $k$, then $\mathbf{B} \in \mathbb{R}^{d \times r}$, $\mathbf{C} \in \mathbb{R}^{r \times k}$, and $r \ll min(d, k)$. This process can be expressed as:

\begin{equation}
\label{deqn_ex7}
\mathbf{h} = \mathbf{W}_{0}\mathbf{h}_{0} + \frac{\alpha}{r}\mathbf{B}\mathbf{C}\mathbf{h}_{0},
\end{equation}

\noindent where $\mathbf{W}_{0}$ denotes the pretrained weight matrix, which contains the model's built-in knowledge, $\alpha$ is a hyperparameter that acts similarly to the learning rate.

We input $Emb(\mathbf{X})$ into the LLM with LoRA and get the output $Emb(\mathbf{H}) \in \mathbb{R}^{N \times D}$. This process can be represented as:

\begin{equation}
\label{deqn_ex8}
\mathbf{H} = LLM(Emb(\mathbf{X})),
\end{equation}

\noindent where $LLM(.)$ denotes the pretrained LLM with LoRA.

Finally, to ensure that the final output matches the desired shape, the predictions are output through a linear layer. This process can be represented as:

\begin{equation}
\label{deqn_ex9}
\mathbf{Y} = ReLU(Linear_{T'}(\mathbf{H})),
\end{equation}

\noindent where $Linear_{T'}(.)$ denotes a linear layer that maps the input to dimension $T'$.

\subsection{Loss Function}
Considering that there are usually outliers caused by sensor failures and other reasons in traffic datasets, we choose the robust L1 loss, i.e. Mean Absolute Error (MAE), as the loss function of the TPLLM:

\begin{equation}
\label{deqn_ex10}
Loss(\mathbf{y}_{i}, \hat{\mathbf{y}}_{i}) = \left|\mathbf{y}_{i} - \hat{\mathbf{y}}_{i}\right|.
\end{equation}

\section{Experiments}
In this section, we conduct experiments on two traffic spatio-temporal datasets. The experiments include full-sample prediction, few-shot prediction, sensitivity analysis to the rank of LoRA, and ablation experiments, which can verify the effectiveness of the cross-modal knowledge transfer capability and few-shot learning capability of pretrained LLMs for traffic prediction tasks.
\subsection{Datasets}
To validate the performance of the proposed framework, we conducted experiments on two real-world traffic datasets PeMS04 and PeMS08. The PeMS datasets \cite{chen2001freeway} contain traffic data collected by multiple sensors on major roads in California at a frequency of 5 minutes. The datasets contain three features: traffic flow, occupancy, and velocity. Since the cyclical variation of traffic flow over time is the most obvious, we take it as the object of the study. In order to improve the stability of the training process, we first normalize all the data and finally renormalize the predictions of the model output. The description of the two datasets used are shown in Table I.

\begin{table}[!h]
\caption{Datasets Description\label{tab:table1}}
\renewcommand\arraystretch{1.25}
\centering
\resizebox{0.8\linewidth}{!}{
\begin{tabular}{cccc}
\toprule
Dataset & Nodes & Time Steps & Time Range \\
\midrule
PeMS04 & 307 & 16992 & 1/1/2018-2/28/2018 \\
PeMS08 & 170 & 17856 & 7/1/2016-8/31/2016 \\
\bottomrule
\end{tabular}
}
\end{table}

In all experiments, we use 1-hour historical traffic flow data as input, i.e., the input sequence length $T$ is 12, to predict the future traffic flow for 15 minutes, 30 minutes and 1 hour, i.e., the output sequence lengths $T'$ is 3, 6 and 12, respectively.

\subsection{Settings}
We use the GPT-2 \cite{radford2019language} as the base LLM for the TPLLM, corresponding to $D = 768$. By applying LoRA, only about 0.95\% of the parameters in the framework are trainable, which significantly reduces the computational cost. In addition, since cross-modal knowledge transfer capability is prevalent in various types of pretrained LLMs, other models can be used in place of GPT-2 as needed.

The optimizer of the TPLLM is set as Adam, and the hyperparameters are shown in Table II.

\begin{table}[!t]
\caption{Hyperparameters of the TPLLM\label{tab:table2}}
\renewcommand\arraystretch{1.25}
\centering
\resizebox{0.8\linewidth}{!}{
\begin{tabular}{cc}
\toprule
Hyperparameter & Value \\
\midrule
batch size & 16 \\
epochs & 500 \\
initial learning rate & 0.001 \\
learning rate decay rate & 0.5/100 epochs \\
$F$ & 64 \\
$r$ & 4/8/16/32/48/64 \\
$\alpha$ & 32 \\
dropout of LoRA & 0.1 \\
\bottomrule
\end{tabular}
}
\end{table}

We constructed the TPLLM based on Python 3.10 and PyTorch 2.0.1. All experiments are conducted on a server computer with 4 Intel Xeon Platinum 8375C CPUs, 6 NVIDIA GeForce RTX 4090 GPUs, and 256 GB of RAM.

\subsection{Evaluation Metrics}
We evaluate the performance of the proposed framework using three common metrics: Mean Absolute Error (MAE), Root Mean Square Error (RMSE), and Mean Absolute Percentage Error (MAPE). For a single prediction of the framework, the process of calculating the evaluation metrics is as follows:

\begin{equation}
\label{deqn_ex11}
MAE(\mathbf{y}_{i}, \hat{\mathbf{y}}_{i}) = \frac{1}{N}\sum_{i=1}^{N}\left|\mathbf{y}_{i} - \hat{\mathbf{y}}_{i}\right|,
\end{equation}

\begin{equation}
\label{deqn_ex12}
RMSE(\mathbf{y}_{i}, \hat{\mathbf{y}}_{i}) = \sqrt{\frac{1}{N}\sum_{i=1}^{N}(\mathbf{y}_{i} - \hat{\mathbf{y}}_{i})^{2}},
\end{equation}

\begin{equation}
\label{deqn_ex13}
MAPE(\mathbf{y}_{i}, \hat{\mathbf{y}}_{i}) = \frac{1}{N}\sum_{i=1}^{N}\frac{\left|\mathbf{y}_{i} - \hat{\mathbf{y}}_{i}\right|}{\mathbf{y}_{i}},
\end{equation}
where $\hat{\mathbf{y}}_{i}$ is the predicted value of traffic flow for the i-th sensor, $\mathbf{y}_{i}$ is the true value of traffic flow for the i-th sensor.

\subsection{Baselines}
The TPLLM is compared with several baselines. The baselines are described in detail below:

\begin{itemize}
\item{LSTM \cite{hochreiter1997long}: Long Short-Term Memory neural networks, a model designed to solve the long-term dependency problem of general RNNs, contains multiple gated units. We constructed an LSTM model with two hidden layers and one fully connected layer.}
\item{STGCN \cite{yu2017spatio}: Spatio-Temporal Graph Convolutional Networks, a model consisting of multiple spatio-temporal convolutional modules, each containing two temporal gated convolutional layers and one spatial graph convolutional layer.}
\item{ASTGCN \cite{guo2019attention}: Attention based Spatial-Temporal Graph Convolutional Networks, a spatio-temporal attention-based model, uses graph convolution and temporal convolution to extract features by combining attention mechanisms.}
\item{STSGCN \cite{song2020spatial}: Spatial-Temporal Synchronous Graph Convolutional Networks, a model that applies graph convolution for synchronized feature extraction by connecting the same nodes at adjacent times to construct a local spatio-temporal graph.}
\end{itemize}

\subsection{Full-sample Prediction}
In full-sample prediction experiments, we explore the effectiveness of pretrained LLMs for a standard traffic prediction task. We divided the datasets into training, validation, and testing sets in chronological order, with the proportions of 60\%, 20\%, and 20\%, respectively. We conducted experiments with different hyperparameter $r$ and took the best results for comparison. A discussion of the hyperparameter $r$ is given in Section \textit{H}.

Table III shows the performance of the TPLLM and other baselines on both datasets, including the evaluation metrics for the next 15-minute, 30-minute, and 1-hour predictions and the average of the evaluation metrics for the predictions of all the time steps within 1 hour.

\begin{table*}[!t]
\caption{Results of Full-sample Prediction\label{tab:table3}}
\renewcommand\arraystretch{1.25}
\centering
\resizebox{\textwidth}{!}{
\begin{threeparttable}
\begin{tabular}{cccccccccccccc}
\hline
\multirow{2}{*}{Dataset} & \multirow{2}{*}{Model} & \multicolumn{3}{c}{15 min (T'=3)}                  & \multicolumn{3}{c}{30 min (T'=6)}                  & \multicolumn{3}{c}{60 min (T'=12)}                 & \multicolumn{3}{c}{Average}                        \\ \cline{3-14} 
                         &                        & MAE            & RMSE           & MAPE             & MAE            & RMSE           & MAPE             & MAE            & RMSE           & MAPE             & MAE            & RMSE           & MAPE             \\ \hline
\multirow{5}{*}{PeMS04}  & LSTM                   & 29.10          & 45.45          & 21.07\%          & 27.25          & 43.66          & 19.26\%          & 32.28          & 49.57          & 22.45\%          & 29.23          & 45.84          & 20.76\%          \\
                         & STGCN                  & 23.56          & 38.16          & 16.07\%          & 25.31          & 40.39          & 17.14\%          & 30.29          & 46.70          & 20.74\%          & 25.92          & 41.28          & 17.66\%          \\
                         & ASTGCN                 & 19.68          & 31.10          & 12.85\%          & 21.52          & 33.89          & 13.99\%          & 25.78          & 39.99          & 16.94\%          & 21.83          & 34.48          & 14.25\%          \\
                         & STSGCN                 & 20.09          & 32.09          & 14.70\%          & 21.60          & 34.18          & 15.52\%          & 24.83          & 38.51          & 18.35\%          & 21.78          & 34.45          & 15.68\%          \\ \cline{2-14} 
                         & TPLLM                  & \textbf{18.39} & \textbf{30.06} & \textbf{12.04\%} & \textbf{19.43} & \textbf{31.76} & \textbf{12.67\%} & \textbf{21.49} & \textbf{34.81} & \textbf{14.20\%} & \textbf{19.53} & \textbf{31.93} & \textbf{12.81\%} \\ \hline
\multirow{5}{*}{PeMS08}  & LSTM                   & 24.16          & 37.70          & 18.52\%          & 22.60          & 36.16          & 17.18\%          & 26.94          & 41.05          & 20.07\%          & 24.29          & 37.99          & 18.42\%          \\
                         & STGCN                  & 20.12          & 30.58          & 13.06\%          & 21.24          & 32.28          & 13.72\%          & 24.78          & 37.10          & 15.86\%          & 21.69          & 32.92          & 14.01\%          \\
                         & ASTGCN                 & 16.19          & 25.01          & 10.37\%          & 18.13          & 27.93          & 11.51\%          & 22.16          & 33.53          & 13.82\%          & 18.33          & 28.30          & 11.64\%          \\
                         & STSGCN                 & 16.26          & 25.07          & 10.61\%          & 17.44          & 27.09          & 12.17\%          & 19.59          & 30.41          & 12.69\%          & 17.56          & 27.18          & 12.08\%          \\ \cline{2-14} 
                         & TPLLM                  & \textbf{14.29} & \textbf{22.96} & \textbf{9.15\%}  & \textbf{15.43} & \textbf{25.34} & \textbf{9.83\%}  & \textbf{17.22} & \textbf{28.43} & \textbf{11.05\%} & \textbf{15.45} & \textbf{25.35} & \textbf{9.88\%}  \\ \hline
\end{tabular}
\begin{tablenotes}
    \footnotesize
    \item The best results are marked in bold.
\end{tablenotes}
\end{threeparttable}
}
\end{table*}

Based on the experimental results, the TPLLM achieves accurate prediction of future traffic flow for 15 minutes, 30 minutes, and 1 hour on both datasets and performs better than baselines on all metrics. This demonstrates that the prior knowledge of the pretrained LLM can be used to analyze the complex spatio-temporal dependencies in traffic data, and its cross-modal knowledge transfer capability is effective for the traffic prediction task.

In all baselines, LSTM can only capture the temporal dependencies of traffic flow, thus its prediction accuracy is poor. The remaining baselines are methods that consider spatio-temporal dependencies and have better prediction accuracy. The TPLLM not only takes into account the spatio-temporal dependencies of traffic flow but also utilizes the prior knowledge of pretrained LLMs to achieve excellent prediction accuracy.

\subsection{Few-shot Prediction}
Due to the high cost of data collection and preservation, few-shot traffic prediction tasks represent common real-world situations. In order to evaluate the advantages that the few-shot learning capability of pretrained LLMs brings to traffic prediction tasks, we conducted experiments in a few-shot setting.

Similar to the full-sample experiments, we divide the datasets into training, validation, and testing sets in chronological order. However, the size of the training set in this experiment is only 10\% of the full-sample experiment, and the validation and test sets are the same as the full-sample experiment, so they account for 6\%, 20\%, and 20\% of the dataset, respectively. We conducted experiments with different hyperparameter r and took the best results for comparison. A discussion of the hyperparameter $r$ is given in Section \textit{H}.

Table IV shows the few-shot performance of the TPLLM and other baselines on both datasets, including the evaluation metrics for the next 15-minute, 30-minute, and 1-hour predictions and the average of the evaluation metrics for the predictions of all the time steps within 1 hour. In addition, Table IV contains the changes of evaluation metrics for the few-shot prediction relative to the full-sample prediction.

\begin{table*}[!t]
\caption{Results of Few-shot Prediction\label{tab:table4}}
\renewcommand\arraystretch{1.23}
\centering
\resizebox{\textwidth}{!}{
\begin{threeparttable}
\begin{tabular}{cccccccccccccc}
\hline
\multirow{2}{*}{Dataset}                                                      & \multirow{2}{*}{Model}  & \multicolumn{3}{c}{15 min (T'=3)}                        & \multicolumn{3}{c}{30 min (T'=6)}                        & \multicolumn{3}{c}{60 min (T'=12)}                       & \multicolumn{3}{c}{Average}                              \\ \cline{3-14} 
                                                                              &                         & MAE              & RMSE             & MAPE               & MAE              & RMSE             & MAPE               & MAE              & RMSE             & MAPE               & MAE              & RMSE             & MAPE               \\ \hline
\multirow{10}{*}{\begin{tabular}[c]{@{}c@{}}PeMS04\\ (Few-shot)\end{tabular}} & \multirow{2}{*}{LSTM}   & 36.21            & 62.26            & 23.85\%            & 34.63            & 61.21            & 21.95\%            & 39.33            & 65.28            & 25.50\%            & 36.42            & 62.62            & 23.59\%            \\
                                                                              &                         & (7.11↑)          & (16.81↑)         & (0.0278↑)          & (7.38↑)          & (17.55↑)         & (0.0269↑)          & (7.05↑)          & (15.71↑)         & \textbf{(0.0305↑)} & (7.19↑)          & (16.78↑)         & (0.0283↑)          \\
                                                                              & \multirow{2}{*}{STGCN}  & 28.20            & 44.57            & 19.16\%            & 30.05            & 46.95            & 20.35\%            & 35.19            & 53.50            & 24.07\%            & 30.67            & 47.85            & 20.85\%            \\
                                                                              &                         & (4.64↑)          & (6.41↑)          & (0.0309↑)          & (4.74↑)          & (6.56↑)          & (0.0321↑)          & \textbf{(4.90↑)} & \textbf{(6.80↑)} & (0.0333↑)          & (4.75↑)          & (6.57↑)          & (0.0319↑)          \\
                                                                              & \multirow{2}{*}{ASTGCN} & 24.12            & 37.39            & 15.55\%            & 26.65            & 40.93            & 17.14\%            & 32.10            & 48.36            & 20.81\%            & 27.12            & 41.76            & 17.52\%            \\
                                                                              &                         & (4.44↑)          & (6.29↑)          & (0.0270↑)          & (5.13↑)          & (7.04↑)          & (0.0315↑)          & (6.32↑)          & (8.37↑)          & (0.0387↑)          & (5.29↑)          & (7.28↑)          & (0.0327↑)          \\
                                                                              & \multirow{2}{*}{STSGCN} & 24.37            & 37.97            & 16.86\%            & 26.59            & 40.90            & 19.14\%            & 31.49            & 47.45            & 21.40\%            & 27.04            & 41.52            & 19.18\%            \\
                                                                              &                         & (4.28↑)          & (5.88↑)          & (0.0216↑)          & (4.99↑)          & (6.72↑)          & (0.0362↑)          & (6.66↑)          & (8.94↑)          & \textbf{(0.0305↑)} & (5.26↑)          & (7.07↑)          & (0.0350↑)          \\ \cline{2-14} 
                                                                              & \multirow{2}{*}{TPLLM}  & \textbf{20.51}   & \textbf{32.62}   & \textbf{13.55\%}   & \textbf{23.27}   & \textbf{36.55}   & \textbf{15.23\%}   & \textbf{28.87}   & \textbf{44.41}   & \textbf{19.00\%}   & \textbf{23.68}   & \textbf{37.38}   & \textbf{15.57\%}   \\
                                                                              &                         & \textbf{(2.12↑)} & \textbf{(2.56↑)} & \textbf{(0.0151↑)} & \textbf{(3.84↑)} & \textbf{(4.79↑)} & \textbf{(0.0256↑)} & (7.38↑)          & (9.60↑)          & (0.0480↑)          & \textbf{(4.15↑)} & \textbf{(5.45↑)} & \textbf{(0.0276↑)} \\ \hline
\multirow{10}{*}{\begin{tabular}[c]{@{}c@{}}PeMS08\\ (Few-shot)\end{tabular}} & \multirow{2}{*}{LSTM}   & 34.46            & 60.15            & 27.49\%            & 33.25            & 59.50            & 26.24\%            & 36.88            & 62.65            & 28.42\%            & 34.62            & 60.54            & 27.28\%            \\
                                                                              &                         & (10.30↑)         & (22.45↑)         & (0.0897↑)          & (10.65↑)         & (23.34↑)         & (0.0906↑)          & (9.94↑)          & (21.60↑)         & (0.0835↑)          & (10.33↑)         & (22.55↑)         & (0.0886↑)          \\
                                                                              & \multirow{2}{*}{STGCN}  & 25.71            & 41.28            & 15.70\%            & 26.86            & 42.81            & 16.33\%            & 30.23            & 46.93            & 18.47\%            & 27.27            & 43.32            & 16.65\%            \\
                                                                              &                         & (5.59↑)          & (10.70↑)         & (0.0264↑)          & (5.62↑)          & (10.53↑)         & (0.0261↑)          & (5.45↑)          & (9.83↑)          & \textbf{(0.0261↑)} & (5.58↑)          & (10.40↑)         & (0.0264↑)          \\
                                                                              & \multirow{2}{*}{ASTGCN} & 19.04            & 29.03            & 11.86\%            & 22.03            & 33.60            & 13.15\%            & 28.17            & 42.82            & 16.74\%            & 22.47            & 34.66            & 13.56\%            \\
                                                                              &                         & (2.85↑)          & (4.02↑)          & (0.0149↑)          & (3.90↑)          & (5.67↑)          & (0.0164↑)          & (6.01↑)          & (9.29↑)          & (0.0292↑)          & (4.14↑)          & (6.36↑)          & (0.0192↑)          \\
                                                                              & \multirow{2}{*}{STSGCN} & 21.98            & 36.16            & 13.71\%            & 23.56            & 38.32            & 14.35\%            & 27.04            & 43.12            & 16.30\%            & 23.88            & 38.73            & 14.87\%            \\
                                                                              &                         & (5.72↑)          & (11.09↑)         & (0.0310↑)          & (6.12↑)          & (11.23↑)         & (0.0218↑)          & (7.45↑)          & (12.71↑)         & (0.0361↑)          & (6.32↑)          & (11.55↑)         & (0.0279↑)          \\ \cline{2-14} 
                                                                              & \multirow{2}{*}{TPLLM}  & \textbf{15.85}   & \textbf{24.75}   & \textbf{10.13\%}   & \textbf{17.76}   & \textbf{27.90}   & \textbf{11.42\%}   & \textbf{21.82}   & \textbf{33.95}   & \textbf{14.09\%}   & \textbf{18.09}   & \textbf{28.51}   & \textbf{11.63\%}   \\
                                                                              &                         & \textbf{(1.56↑)} & \textbf{(1.79↑)} & \textbf{(0.0098↑)} & \textbf{(2.33↑)} & \textbf{(2.56↑)} & \textbf{(0.0159↑)} & \textbf{(4.60↑)} & \textbf{(5.52↑)} & (0.0304↑)          & \textbf{(2.64↑)} & \textbf{(3.16↑)} & \textbf{(0.0175↑)} \\ \hline
\end{tabular}
\begin{tablenotes}
    \footnotesize
    \item The best results are marked in bold.
    \item ↑ denotes the extent to which the evaluation metrics have risen relative to the full-sample prediction.
\end{tablenotes}
\end{threeparttable}
}
\end{table*}

Based on the experimental results, the TPLLM outperforms the baselines for all metrics on both datasets, and even some of the metrics outperform the baselines in the full-sample experiment. Further, most of the metrics of the TPLLM are the least changed relative to the full-sample experiment. This demonstrates that the TPLLM is able to make predictions with high accuracy in the lack of training data and that the few-shot learning capability of pretrained LLMs is effective for traffic prediction tasks.

\subsection{Visualization of Full-sample and Few-shot Prediction}
\begin{figure}[!t]
\centering
\subfloat[]{\includegraphics[width=0.9\linewidth]{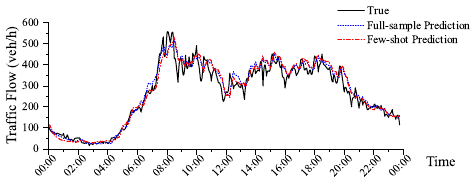}%
\label{fig_6-1}}\\
\subfloat[]{\includegraphics[width=0.9\linewidth]{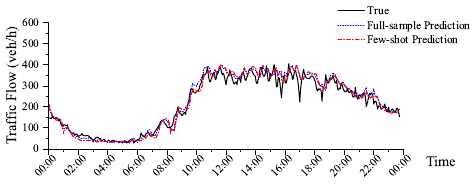}%
\label{fig_6-2}}\\
\subfloat[]{\includegraphics[width=0.9\linewidth]{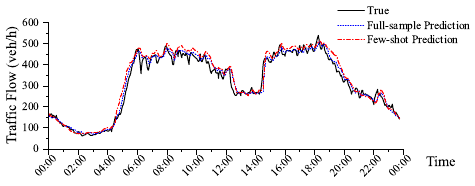}%
\label{fig_6-3}}\\
\subfloat[]{\includegraphics[width=0.9\linewidth]{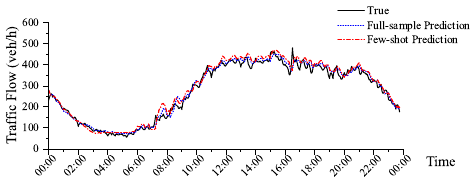}%
\label{fig_6-4}}
\caption{Visualization charts of experimental results. (a) Predictions of a working day in PeMS04. (b) Predictions of a weekend day in PeMS04. (c) Predictions of a working day in PeMS08. (d) Predictions of a weekend day in PeMS08.
}
\label{fig_6}
\end{figure}

For each of the two datasets, we randomly selected a single node and two random days in the test set (one day taken from weekdays and one from weekends). Accordingly, visualization charts of full-sample prediction and few-shot prediction are drawn to observe the results more intuitively, as shown in Fig.~\ref{fig_6}.

It can be seen that the TPLLM has good accuracy in full-sample prediction, whether it is the high traffic flow caused by the morning and evening peaks on weekdays or the traffic flow that is smooth all day long on rest days. As for the few-shot prediction, although the accuracy is a little bit worse, it can also make a good prediction of the changing trend of traffic flow, which can meet the needs of urban traffic management.

\subsection{Ablation Study}
In order to validate the effectiveness of each module in the TPLLM, we remove the graph embedding layer, sequence embedding layer, and LoRA from the framework, respectively. Subsequently, we conducted experiments on both datasets while keeping other hyperparameters unchanged.

Table V shows the full-sample prediction and few-shot prediction performance (average of the evaluation metrics for the predictions of all the time steps within 1 hour) of the TPLLM and its degradation models on the two datasets.

\begin{table}[!t]
\caption{Results of Ablation Experiment\label{tab:table5}}
\renewcommand\arraystretch{1.25}
\centering
\resizebox{0.8\linewidth}{!}{
\begin{threeparttable}
\begin{tabular}{ccccc}
\hline
Dataset                                                                      & Model   & MAE            & RMSE           & MAPE             \\ \hline
\multirow{4}{*}{PeMS04}                                                      & No SE   & 21.22          & 35.49          & 13.86\%          \\
                                                                             & No GE   & 20.20          & 32.54          & 13.82\%          \\
                                                                             & No LoRA & 23.76          & 37.51          & 16.15\%          \\ \cline{2-5} 
                                                                             & TPLLM   & \textbf{19.53} & \textbf{31.93} & \textbf{12.81\%} \\ \hline
\multirow{4}{*}{\begin{tabular}[c]{@{}c@{}}PeMS04\\ (Few-shot)\end{tabular}} & No SE   & 30.44          & 48.89          & 20.52\%          \\
                                                                             & No GE   & 23.98          & 37.86          & 15.73\%          \\
                                                                             & No LoRA & 24.64          & 38.43          & 17.69\%          \\ \cline{2-5} 
                                                                             & TPLLM   & \textbf{23.68} & \textbf{37.38} & \textbf{15.57\%} \\ \hline
\multirow{4}{*}{PeMS08}                                                      & No SE   & 19.45          & 31.54          & 12.43\%          \\
                                                                             & No GE   & 15.98          & 25.75          & 10.26\%          \\
                                                                             & No LoRA & 19.16          & 30.33          & 12.00\%          \\ \cline{2-5} 
                                                                             & TPLLM   & \textbf{15.45} & \textbf{25.35} & \textbf{9.88\%}  \\ \hline
\multirow{4}{*}{\begin{tabular}[c]{@{}c@{}}PeMS08\\ (Few-shot)\end{tabular}} & No SE   & 27.86          & 48.39          & 18.16\%          \\
                                                                             & No GE   & 18.58          & 29.24          & 12.45\%          \\
                                                                             & No LoRA & 20.43          & 31.88          & 14.57\%          \\ \cline{2-5} 
                                                                             & TPLLM   & \textbf{18.09} & \textbf{28.51} & \textbf{11.63\%} \\ \hline
\end{tabular}
\begin{tablenotes}
    \footnotesize
    \item The best results are marked in bold.
\end{tablenotes}
\end{threeparttable}
}
\end{table}

The experimental results show that the original framework outperforms the 3 degenerate models. Therefore, the sequence embedding layer, graph embedding layer, and LoRA all positively affect the prediction performance of the framework.

The framework without the sequence embedding layer performs the worst in all predictions except the full-sample PeMS04, which indicates that the sequence embedding layer has the greatest impact on prediction accuracy. We believe that since the essence of traffic data is still time-series data, the features of the sequence itself are the most important. The sequence embedding layer consisting of 1-D CNNs can effectively extract the features of the sequence itself, so this module is indispensable.

The framework without the graph embedding layer shows decreased accuracy in all the experiments, indicating that the application of GCNs helps to effectively extract spatial features of traffic data, and the spatial features help to improve the accuracy of traffic prediction in complex road networks.

The framework without LoRA shows decreased accuracy in all experiments, indicating that LoRA does enable the pretrained LLM to learn new knowledge from the features extracted from the embedding layer that is beneficial for traffic prediction tasks. In addition, in the framework with LoRA removed, the parameters of the pretrained LLM are completely frozen and the LLM cannot learn any new knowledge. In this case, the output prediction does not have a great loss of accuracy, which reflects some extent the zero-shot learning capability \cite{brown2020language} of the pretrained LLM, and its prior knowledge is effective for traffic prediction tasks.

\subsection{Sensitivity to rank of LoRA}
The rank $r$ is an important hyperparameter in LoRA that reflects the amount of information that the decomposition matrix $\mathbf{B}$ and $\mathbf{C}$ can contain. A small $r$ may make the size of the decomposition matrix insufficient to fully accommodate the new knowledge, and a large $r$ may introduce too much redundant information and create noise.

\begin{figure}[!t]
\centering
\subfloat[]{\includegraphics[width=0.9\linewidth]{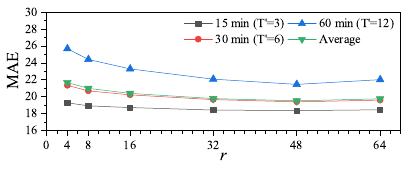}%
\label{fig_7-1}}\\
\subfloat[]{\includegraphics[width=0.9\linewidth]{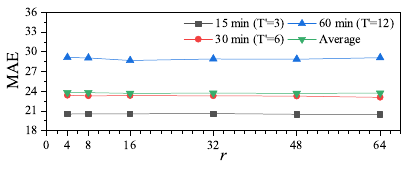}%
\label{fig_7-2}}\\
\subfloat[]{\includegraphics[width=0.9\linewidth]{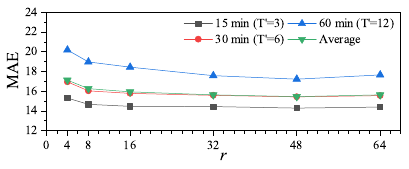}%
\label{fig_7-3}}\\
\subfloat[]{\includegraphics[width=0.9\linewidth]{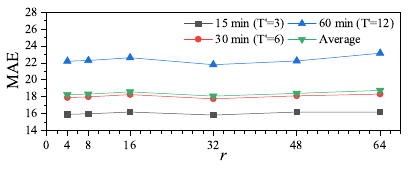}%
\label{fig_7-4}}
\caption{Mean absolute error for experiments with different r. (a) Full-sample prediction of PeMS04. (b) Few-shot prediction of PeMS04. (c) Full-sample prediction of PeMS08. (d) Few-shot prediction of PeMS08.}
\label{fig_7}
\end{figure}

In order to figure out the effect of $r$, we applied different $r$ for full-sample and few-shot prediction, and the results are shown in Fig.~\ref{fig_7}.

For both predictions of PeMS04 and the full-sample prediction of PeMS08, the optimal $r$ is 48; while for the small-sample prediction of PeMS08, the optimal $r$ is 32. This may be due to the fact that the amount of knowledge learned by the pretrained LLM in the few-shot prediction of PeMS08 is smaller than in other prediction experiments, and thus a smaller $r$ is more appropriate.

However, in summary, the MAE curves of all the prediction results are relatively flat, which indicates that $r$ has little effect on the accuracy. Therefore, a smaller $r$ can be chosen to reduce the size of the learnable matrices $\mathbf{B}$ and $\mathbf{C}$ in practical applications, thus saving computational resources and obtaining sufficiently effective predictions.

\section{Conclusions}
Achieving accurate traffic predictions in areas with limited historical traffic data is a challenging task. The aim of this study is to introduce pretrained LLMs for traffic prediction tasks and to utilize the few-shot learning capability of LLMs to overcome the challenges caused by the lack of data. Therefore, we propose TPLLM, a framework for traffic prediction based on pretrained LLMs. The main conclusions are summarized as follows:

\begin{itemize}
\item{The TPLLM outperforms the other baselines in full-sample prediction, indicating that the embedding layer and the pretrained LLM have advantages in analyzing spatio-temporal dependencies in traffic data.}
\item{The TPLLM outperforms the other baselines in few-shot prediction (10\% training set) with less degradation of metrics, which indicates that the few-shot learning capability of pretrained LLM is beneficial for small-sample traffic prediction tasks that are close to realistic scenarios.}
\item{The rank of LoRA has less effect on the accuracy of the prediction results, therefore, a smaller rank can be chosen in practical applications to obtain sufficiently accurate predictions while saving computational resources.}
\item{The experimental results of the ablation study show that the sequence embedding layer, the graph embedding layer, and LoRA in the TPLLM all have a positive effect on the prediction accuracy.}
\end{itemize}

In summary, the TPLLM is a novel traffic prediction framework that can effectively capture the spatio-temporal features of graph-structured traffic data and provide accurate predictions based on pretrained LLMs.

For future works, we expect to design embeddings to improve prediction accuracy by considering more factors that affect traffic. In addition, we will further explore PEFT techniques that are more applicable to spatio-temporal prediction tasks and try to find an interpretable knowledge learning pattern of LLMs.


\bibliographystyle{IEEEtran}
\small\bibliography{reference}

\newpage

 




\vfill

\end{document}